\documentclass{article}

 \usepackage[main, final]{neurips_2026}

\usepackage[utf8]{inputenc} 
\usepackage[T1]{fontenc}    
\usepackage{hyperref}       
\usepackage{url}            
\usepackage{booktabs}       
\usepackage{amsfonts}       
\usepackage{nicefrac}       
\usepackage{microtype}      
\usepackage{xcolor}         

\usepackage{multirow}
\usepackage{enumitem}
\usepackage{makecell}
\usepackage{amsmath}
\usepackage{amssymb}
\usepackage{graphicx}
\usepackage{subcaption}
\usepackage{gensymb}

\def\PaperTitle {HOPE: Hand-Object Pressure Estimation\\from Monocular Videos}

\newcommand{\fref}[1]{Fig.~\ref{#1}}

\newcommand{\textblock}[1]{\noindent\textbf{#1}}

\title{\PaperTitle}

\author{%
  Subin Jeon\textsuperscript{1} \quad Byungjun Kim\textsuperscript{1} \quad Hanbyul Joo\textsuperscript{1,2}\\
  \textsuperscript{1}Seoul National University \quad \textsuperscript{2}RLWRLD \\
  \texttt{\{subinjeon, byungjun.kim, hbjoo\}@snu.ac.kr} \\
  \\[0.3em]
  {\tt\small \href{https://subin6.github.io/page-hope/}{\color{magenta}{https://subin6.github.io/page-hope/}}}
}
\vspace{-5mm}

\begin{document}

\maketitle

\begin{abstract}
Estimating physical pressure from vision is essential for understanding contact-rich hand-object interaction.
However, prior vision-based pressure estimation methods are largely limited to planar surfaces and single image input, making them difficult to apply to dynamic hand-object interaction with diverse objects.
We instead formulate pressure estimation as a hand-centric video prediction problem with monocular video as input.
This formulation predicts temporally evolving per-vertex normal pressure and contact directly on the hand mesh, yielding a unified output space independent of object shape and sensor layout.
Building on this formulation, we propose \textbf{HOPE}, a framework with two key components. First, we lift tactile-glove pressure, planar-sensor pressure, and distance-based hand-object contact annotations into a shared hand vertex space, allowing bare-hand contact data to regularize pressure learning where labels are unavailable. Second, we introduce a vertex-anchored video transformer that treats each vertex as a persistent token, aggregates visual features and hand pose over time, and uses a contact-gated pressure head to suppress pressure predictions in regions with low contact probability.
Experiments on OpenTouch, PressureVisionDB, and hand-object contact benchmarks validate HOPE across object-pressure, surface-pressure, and contact-supervised HOI settings.
Despite using pressure supervision primarily from gloved-hand videos, HOPE generalizes to bare-hand egocentric and in-the-wild videos, producing joint contact and pressure predictions beyond the scope of contact-only or planar-pressure baselines.

\end{abstract}

\section{Introduction}
Estimating physical pressure from visual observations is a key step toward understanding how humans manipulate the world.
While RGB video reveals where hands and objects are, many downstream tasks require knowing how they physically interact: which regions of the hand are in contact, how contact evolves over time, and where pressure is applied during pressing, grasping, or tool use.
Such information is important for egocentric perception, augmented and virtual reality, robot learning from human demonstrations, and assistive systems that need to reason about physical interaction beyond object identity or hand pose alone.
\begin{figure}[t]
\centering
\includegraphics[trim={0 0 0 0pt}, width=1\linewidth]{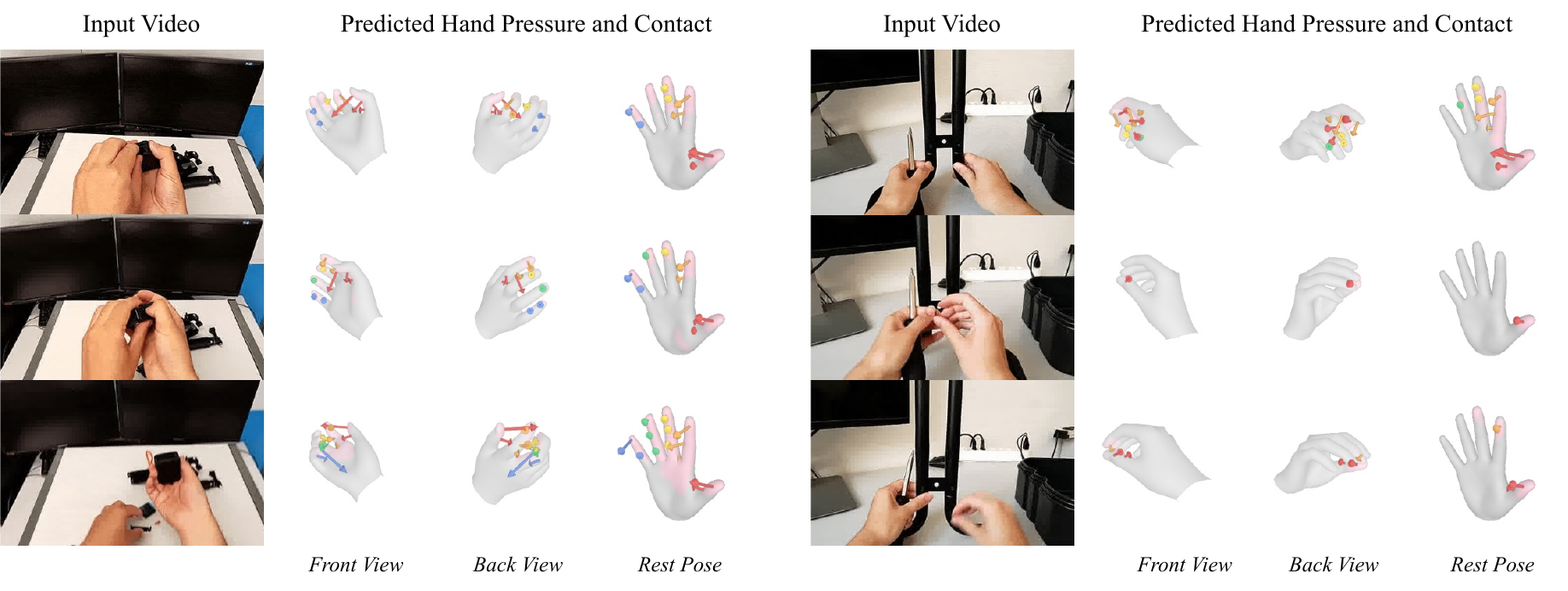}
    \caption{From RGB video to hand-object pressure and contact. \textbf{HOPE} predicts vertex-level contact and pressure on the hand mesh without tactile sensing at test time.
Pink vertices denote contact, and arrows indicate the averaged pressure computed from the vertex-level pressure predictions.
}\label{fig:teaser}
\end{figure}

Despite this importance, vision-based pressure estimation has so far been studied mostly under constrained sensing geometries. Prior methods \cite{grady2022pressurevision, grady2024pressurevision++, zhao2025egopressure} have made important progress by estimating pressure from RGB observations in calibrated settings, often using instrumented surfaces where the contact surface is known and pressure can be represented as a 2D planar map.
These formulations provide valuable physical supervision, but they remain tied to a particular sensing surface and a limited range of interactions.
Everyday hand-object interaction is less structured: the hand contacts arbitrary 3D object surfaces, undergoes frequent self- and object-occlusion, and exhibits diverse contact modes such as fingertip pressing, whole-hand grasping, palm support, and tool use.
As a result, generalizing pressure estimation beyond instrumented surfaces to arbitrary hand-object interaction remains difficult.

Recent hand-centric tactile datasets for object manipulation, exemplified by OpenTouch~\cite{song2025opentouch}, mark an important step toward pressure estimation beyond fixed sensing surfaces.
By moving tactile sensing from the tactile sensor pad to the hand, they make it possible to collect pressure annotations while the same hand interacts with many different objects.
However, this benefit comes with a new limitation: the tactile glove that enables measurement also covers the hand.
As a result, many appearance cues used by prior vision-based pressure methods, such as skin deformation, color change, local wrinkles, and fingertip appearance, are no longer directly visible.
The visual domain of gloved hands also differs from the bare hands that appear in everyday manipulation, making it difficult for a model trained only on wearable tactile data to generalize to natural bare-hand interaction.

To overcome these limitations, we learn hand-object pressure from complementary pressure and contact supervision.
Wearable tactile data provides continuous pressure readings during object manipulation, but is sensor-specific and visually tied to gloved hands; 3D HOI data provides diverse bare-hand interactions, but only contact.
We unify these sources by lifting tactile-glove pressure, planar-sensor pressure, and distance-based HOI contact into a shared MANO~\cite{romero2017mano} vertex space, where both pressure and contact are represented as per-vertex fields.
Although contact does not determine pressure magnitude, it provides a structural prior on where pressure can occur and supplies bare-hand interaction supervision that gloved pressure data alone cannot provide.
This unified representation allows metric pressure labels and broad bare-hand contact data to be used jointly for hand-object pressure estimation.

We further propose a vertex-anchored video transformer for hand pressure estimation.
The model represents each MANO vertex as a persistent token, enriches it with visual and hand-pose evidence, and reasons over both the hand surface and time.
This design is motivated by the temporal nature of pressure.
Pressure is not determined by per-frame appearance alone: it changes as the hand approaches an object, establishes contact, loads the object during grasping or holding, and releases it.
In these phases, visually similar contact configurations may correspond to different pressure magnitudes.
Temporal modeling therefore helps disambiguate loading states and propagates pressure cues through short occlusions.
Finally, a contact-gated pressure head couples the two predictions by suppressing pressure in regions with low predicted contact probability.

We evaluate the proposed framework on hand-object pressure estimation with OpenTouch \cite{song2025opentouch}, hand-surface pressure estimation with PressureVisionDB \cite{grady2022pressurevision}, and dense hand contact estimation on hand-object interaction \cite{fan2023arctic,cao2021reconstructing}.
Our experiments show that the proposed formulation enables pressure estimation beyond fixed planar surfaces and benefits from combining pressure supervision with contact-only HOI data.
Furthermore, as illustrated in Fig.~\ref{fig:teaser}, the learned hand-centric representation can be applied to bare-hand videos, producing dense contact and pressure predictions despite pressure supervision being obtained primarily from gloved-hand videos.

Our contributions are threefold.
First, we introduce a hand-centric formulation for visual hand-object pressure estimation, representing pressure and contact as per-vertex fields on the MANO hand mesh.
Second, we unify heterogeneous supervision by lifting tactile-glove pressure, planar-sensor pressure, and contact-only HOI annotations into the same vertex space, allowing bare-hand contact data to serve as a structural prior for pressure estimation.
Third, we propose a vertex-anchored video transformer with contact-gated pressure prediction, enabling temporally informed pressure estimation from monocular RGB video.
\section{Related Work}
\subsection{Vision-based hand pressure estimation.}
Estimating physical pressure from visual observations has recently emerged as a way to perceive contact-rich interaction without instrumenting every object or surface.
PressureVision and its follow-up showed that RGB images contain pressure cues such as skin deformation, color change, pose, and shadows, while EgoPressure moves this problem into the egocentric setting with paired pressure and pose annotations~\cite{grady2022pressurevision,grady2024pressurevision++,zhao2025egopressure}.
These works establish that pressure can be inferred from visual evidence, but their pressure outputs remain tied to planar sensor images, image-space maps, fingertip-level labels, or a particular capture setup.
Among them, EgoPressure \cite{zhao2025egopressure} is the most closely related to ours: both build on off-the-shelf MANO reconstruction to express pressure in a hand-centric space.
EgoPressure, however, estimates bare-hand pressure frame-wise against an instrumented planar surface, whereas HOPE targets diverse hand-object interactions, jointly learns from heterogeneous pressure and contact supervision, and models temporal pressure evolution.

\subsection{Tactile sensing and touch datasets for hand-object interaction.}
Direct tactile sensing provides physical supervision that is difficult to obtain from vision alone.
Physical touch has been captured with instrumented objects or sensing surfaces, wearable tactile gloves, and thermal contact imprints~\cite{pham2018contactforce,grady2022pressurevision,sundaram2019tactileglove,brahmbhatt2019contactdb,brahmbhatt2020contactpose}.
Such setups provide direct force, pressure, or contact evidence, but their labels are naturally expressed in the coordinate system and coverage of the underlying hardware, such as force transducer locations, planar sensels, thermal object surfaces, or glove taxels.
In contrast, many large hand-object interaction datasets provide hand-object pose, geometry, 2D contact, or inferred contact annotations that support contact reasoning, but do not directly measure pressure~\cite{hasson2019learning,shan2020understanding,cao2021reconstructing,taheri2020grab,Chao2021dexycb,hampali2020honnotate,hampali2022keypoint,kwon2021h2o,liu2022hoi4d,yang2022oakink,fan2023arctic,bhatnagar2022behave}.
Most closely related to our setting, OpenTouch~\cite{song2025opentouch} captures in-the-wild egocentric video, full-hand tactile maps, and hand pose with a wearable tactile glove, making full-hand touch available during natural manipulation.
We use these physical signals as complementary supervision, but lift them into a shared MANO vertex representation so that heterogeneous pressure and contact labels can be learned jointly.

\subsection{Hand-object contact and affordance estimation.}
Hand-object interaction has also been widely studied through contact, grasp, and affordance rather than pressure.
Contact cues have been used to annotate grasp regions, refine hand-object poses, generate plausible grasps, and reconstruct in-the-wild hand-object interactions~\cite{brahmbhatt2019contactdb,brahmbhatt2020contactpose,jiang2021contactgrasp,grady2021contactopt,liu2023contactgen,hu2024explicitcontact,liu2025easyhoi,nam2024jointrecon}.
Large-scale interaction datasets and contact-estimation pipelines further show that contact can be derived from tracked geometry across diverse hand-object settings, although these labels are typically binary or proximity-based rather than metric pressure~\cite{hasson2019learning,shan2020understanding,cao2021reconstructing,taheri2020grab,Chao2021dexycb,hampali2020honnotate,hampali2022keypoint,kwon2021h2o,liu2022hoi4d,yang2022oakink,fan2023arctic}.
Dense vertex-level contact has also been studied for human-scene interaction, while HACO focuses on MANO-vertex hand contact from RGB images under severe class and spatial imbalance~\cite{hassan2021populating,huang2022capturing,tripathi2023deco,jung2025haco}.
Beyond static contact labels, HOI generation and affordance models synthesize hand-object motion, contact maps, or object-centric interaction regions from text, visual observations, object geometry, and learned interaction priors~\cite{christen2024diffh2o,cha2024text2hoi,li2024taskhoi,taheri2024grip,xu2023interdiff,ye2022ihoi,ye2023affordancediffusion,ye2024ghop,zhang2024artigrasp,bahl2023affordances,yang2023affordance3d,prakash2025latentact,kim2024coma,kim2025david}.
Among them, LatentAct is particularly close to our representation because it predicts future 3D hand motion and MANO contact-map trajectories, but it still models where contact occurs rather than how much pressure is applied~\cite{prakash2025latentact}.
These contact and affordance models provide important cues about where interaction is likely to occur, but they do not estimate the magnitude of physical loading on the hand.
Our work treats contact as a structural prior for pressure and jointly predicts contact and per-vertex pressure, thereby connecting contact-rich HOI understanding with hand pressure estimation.

\begin{figure}[t]
\centering
\includegraphics[trim={0 0 0 0pt}, width=1\linewidth]{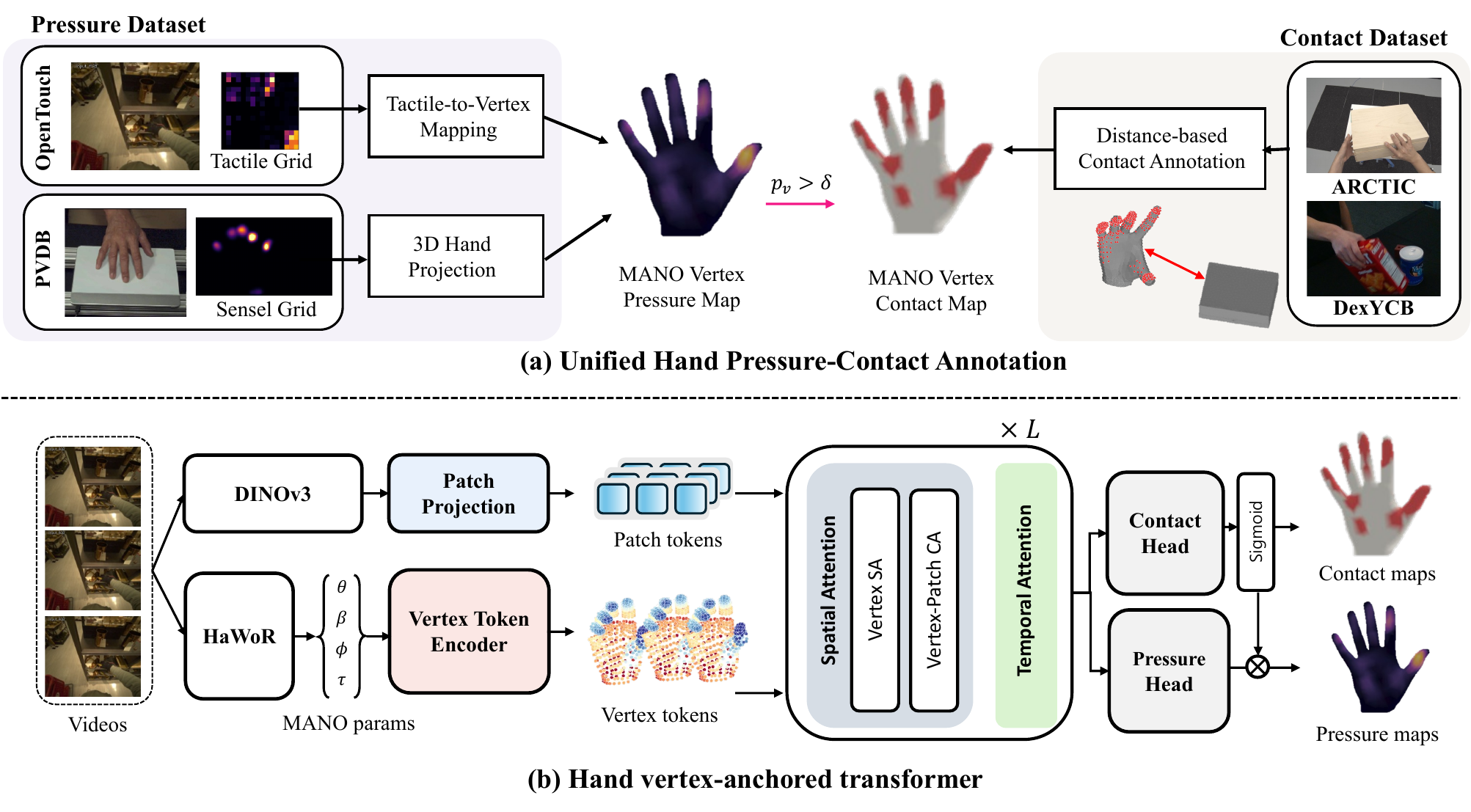}
    \caption{\textbf{Method overview.}
(a)~\textbf{Unified Hand Pressure–Contact Annotation.} Heterogeneous pressure and contact supervision is lifted into a shared MANO vertex space.
(b)~\textbf{VertexFormer.} Image patch tokens and vertex tokens are fused via interleaved spatial and temporal attention.}
   \label{fig:method}
\end{figure}


\newcommand{\methodname}{VertexFormer}      
\newcommand{\OT}{\textsc{OpenTouch}}
\newcommand{\PV}{\textsc{PVDB}}
\newcommand{\hawor}{HaWoR}
\newcommand{\R}{\mathbb{R}}
\newcommand{\Real}{\mathbb{R}}
\newcommand{\Indic}{\mathbf{1}}

\section{Method}
\label{sec:method}
We propose \textbf{HOPE}, a framework for predicting per-vertex contact and pressure on the hand surface from RGB video. Given a clip of $T$ frames $\{I_t\}_{t=1}^{T}$, HOPE predicts per-vertex pressure $\hat{\mathbf{p}}_t \in \Real_{\geq 0}^{V}$ and per-vertex contact $\hat{\mathbf{c}}_t \in [0,1]^{V}$ on the MANO~\cite{romero2017mano} mesh, where $V{=}778$. HOPE consists of two components: (i)~a unified annotation pipeline that lifts tactile-glove pressure, planar-sensor pressure, and distance-based hand-object contact into a shared MANO vertex space (Sec.~\ref{sec:annotation}), and (ii)~\textbf{VertexFormer}, a vertex-anchored video transformer that predicts contact and pressure jointly on the MANO mesh through interleaved spatial and temporal attention with a contact-gated pressure head (Sec.~\ref{sec:architecture}).
Training is performed with a unified objective: samples with pressure labels supervise both per-vertex pressure regression and contact classification, while contact-only HOI samples supervise only the contact head (Sec.~\ref{sec:loss}).

\subsection{Unified pressure--contact annotation on MANO vertex space}
\label{sec:annotation}

Existing pressure datasets provide annotations in sensor-specific coordinate systems---\OT{} \cite{song2025opentouch} in a $16{\times}16$ tactile taxel grid attached to the palmar side of a glove, and PressureVisionDB~(\PV) \cite{grady2022pressurevision} in a 2D sensor image of a planar touchpad. Training a single model across both sources requires a common output space. Moreover, while these datasets provide direct pressure supervision, their sensing setups cover constrained interaction domains. We therefore complement them with distance-based contact annotations from existing HOI datasets \cite{fan2023arctic,jung2025haco}, which provide diverse bare-hand object interactions without pressure labels (\fref{fig:method} (a)).

\textblock{Tactile-to-vertex mapping (\OT).}
\OT{} provides a correspondence $\mathcal{M}: {1,\dots,256} \to 2^{\mathcal{V}}$ between each taxel and the MANO vertices physically covered by its sensing patch, which we project onto the standard $V{=}778$ topology.
Since taxel patches span many vertices (38 on average) and adjacent patches share vertices at their boundaries, we aggregate the taxel pressures $\boldsymbol{\pi} \in \Real_{\geq 0}^{256}$ at each vertex by averaging over its covering taxels,
\begin{equation}
\tilde{p}_v \;=\; \frac{1}{|\mathcal{T}(v)|} \sum_{i \in \mathcal{T}(v)} \pi_i, \qquad \mathcal{T}(v) = \{i : v \in \mathcal{M}(i)\},
\label{eq:ot_mapping}
\end{equation}
yielding a vertex-level pressure field $\mathbf{p}^{\text{OT}} \in \Real_{\geq 0}^{V}$ over the sensor-covered region of the hand.

\textblock{Sensel-to-vertex projection (\PV).}
\PV{} provides dense pressure on a planar sensor pad synchronized with RGB video, but does not include 3D hand annotations. We obtain a posed MANO mesh per frame from an off-the-shelf hand reconstruction model, refined with multi-view fusion. We then project each palmar vertex onto the pad plane using the calibrated camera-to-pad extrinsics and bilinearly sample the sensor pressure map at its $(u,v)$ coordinate, yielding $\mathbf{p}^{\text{PV}} \in \Real_{\geq 0}^{V}$.

\textblock{Pressure and contact supervision.}
\PV{} provides pressure directly in kPa, while \OT{} provides tactile readings that we linearly convert to kPa using its reported pressure range. We use the resulting vertex-level values as pressure supervision. Additionally, we derive binary contact labels by thresholding the pressure, $c_{t,v} \;=\; \Indic\!\left[p_{t,v} > \tau_p\right]$.

\textblock{Distance-based contact from HOI data.}
To complement pressure supervision with diverse bare-hand object interactions, we incorporate contact annotations from existing HOI datasets~(DexYCB \cite{Chao2021dexycb}, ARCTIC \cite{fan2023arctic}) following the protocol of HACO~\cite{jung2025haco}. Given the ground-truth posed hand and object meshes, a vertex is labeled as in contact if its distance to the object surface is below a threshold $\delta_c$,
\begin{equation}
c_{t,v} \;=\; \Indic\!\left[\,\mathrm{dist}\!\left(\mathbf{x}_{t,v}^{\text{hand}},\,\mathcal{S}_t^{\text{obj}}\right) < \delta_c\right],
\label{eq:dist_contact}
\end{equation}
where $\mathbf{x}_{t,v}^{\text{hand}}$ is the world-space coordinate of vertex~$v$ at frame~$t$ and $\mathcal{S}_t^{\text{obj}}$ is the object surface. These annotations provide complementary contact supervision without requiring pressure measurements.
The final training set is the union of pressure samples (with $p_t$ and derived $c_t$) from OpenTouch/PVDB and contact-only samples (with $c_t$ alone) from HOI datasets, all expressed in the same $V$-dimensional MANO vertex space.

\subsection{\methodname: a vertex-anchored transformer}
\label{sec:architecture}
\methodname{} (\fref{fig:method} (b)) tokenizes the input along two complementary axes---image patches and hand vertices---and reasons over both with interleaved spatial and temporal attention. Throughout the network, vertex tokens carry the prediction state and image patches serve as a key--value memory; we refer to this as a \emph{vertex-anchored} design.

\textblock{Patch and vertex tokens.}
Each frame is encoded by a frozen ViT image encoder~\cite{simeoni2025dinov3} into a grid of patch features, which are linearly projected to dimension $d$ to form patch tokens $\mathbf{F}t \in \Real^{N_p \times d}$.
In parallel, an off-the-shelf 3D hand reconstruction model estimates per-frame MANO parameters $(\theta_t, \beta_t, \phi_t, \tau_t)$, where $\theta_t$ and $\beta_t$ denote pose and shape, and $\phi_t$ and $\tau_t$ denote global orientation and camera-relative translation, respectively.
In our implementation, we use the camera-space MANO estimates from HaWoR~\cite{zhang2025hawor}, prior to its world-space reconstruction.
To support fine-grained pressure prediction directly on the hand surface, we embed the posed MANO vertices as persistent vertex tokens. We first obtain hand-local vertex coordinates $\tilde{\mathbf{V}}t \in \mathbb{R}^{V \times 3}$ from the estimated pose and shape, and construct each vertex token as
\begin{equation}
\mathbf{z}_{t,v} \;=\; \mathbf{e}_v \;+\; W_v\,\tilde{\mathbf{V}}_{t,v} \;+\; W_\phi\,\phi_t \;+\; W_\tau\,\tau_t,
\label{eq:vertex_token}
\end{equation}
where $\mathbf{e}_v$ is a learned vertex-identity embedding, and $W_v$, $W_\phi$, and $W_\tau$ are learned projections of the hand-local vertex coordinate, global orientation, and translation, respectively, into the token dimension $d$. The identity embedding maintains a fixed correspondence between tokens and MANO vertices, while separating hand-local geometry from global orientation and translation allows the model to reason separately about hand articulation, viewpoint, and camera-relative position.

\textblock{Spatial vertex--patch attention.}
Each transformer block applies two attention operations over the vertex tokens. \emph{Vertex self-attention}~(Vertex-SA) lets vertices exchange information along the hand topology, propagating contact signals from observable regions to occluded ones. \emph{Vertex-to-patch cross-attention}~(Vertex-Patch CA) lets each vertex token attend to the full patch grid, aggregating visual evidence from the image. Together these realize the vertex-anchored design: vertices are queries and remain the carriers of state, while image features serve as a key--value memory.

\textblock{Temporal attention.}
A subsequent temporal attention layer attends over the $T$ frames per vertex index, sharing information across time at fixed vertex identity. This aggregates temporal evidence for pressure prediction and helps disambiguate contact during brief occlusions. The spatial--temporal block is repeated $L$ times.

\textblock{Contact-gated pressure prediction.}
Two MLP heads take the final vertex tokens as input. The contact head outputs a logit per vertex, passed through a sigmoid to produce $\hat{\mathbf{c}}_t$. The pressure head outputs a non-negative pressure value per vertex, $\tilde{\mathbf{p}}_t$. We then gate the pressure prediction by the contact probability,
\begin{equation}
\hat{\mathbf{p}}_t \;=\; \hat{\mathbf{c}}_t \,\odot\, \tilde{\mathbf{p}}_t,
\label{eq:gating}
\end{equation}
where $\odot$ denotes element-wise product over vertices. 
The gate suppresses pressure where the predicted contact probability is low, allowing contact-only HOI supervision to guide pressure localization.

\subsection{Training objective}
\label{sec:loss}

Let $\Omega_t^p \subseteq \{1,\dots,V\}$ and $\Omega_t^c$ denote the sets of vertices with valid pressure and contact annotations, respectively.
We normalize pressure using a shared reference scale $s=50\,\mathrm{kPa}$ (the nominal full-scale range of OpenTouch) for both \OT{} and \PV{}, defining $q_{t,v}=p_{t,v}/s$ and $\hat{q}_{t,v}=\hat{p}_{t,v}/s$.
We supervise contact with binary cross-entropy and pressure with mean squared error (MSE):
\begin{equation}
\mathcal{L}_{\text{contact}} =
\frac{1}{|\Omega_t^c|}
\sum_{v \in \Omega_t^c}
\mathrm{BCE}(\hat{c}_{t,v}, c_{t,v}),
\qquad
\mathcal{L}_{\text{pressure}} =
\frac{1}{|\Omega_t^p|}
\sum_{v \in \Omega_t^p}
(\hat{q}_{t,v}-q_{t,v})^2.
\label{eq:training_losses}
\end{equation}
The total objective is
$\mathcal{L} =
\lambda_c\mathcal{L}_{\text{contact}}
+
\lambda_p\mathcal{L}_{\text{pressure}},
$
where the pressure term is omitted for contact-only HOI samples.
This unified loss enables joint training over pressure and contact data despite their heterogeneous supervision coverage.

\begin{figure}[t]
\centering
\includegraphics[trim={0 0 0 0pt}, width=1\linewidth]{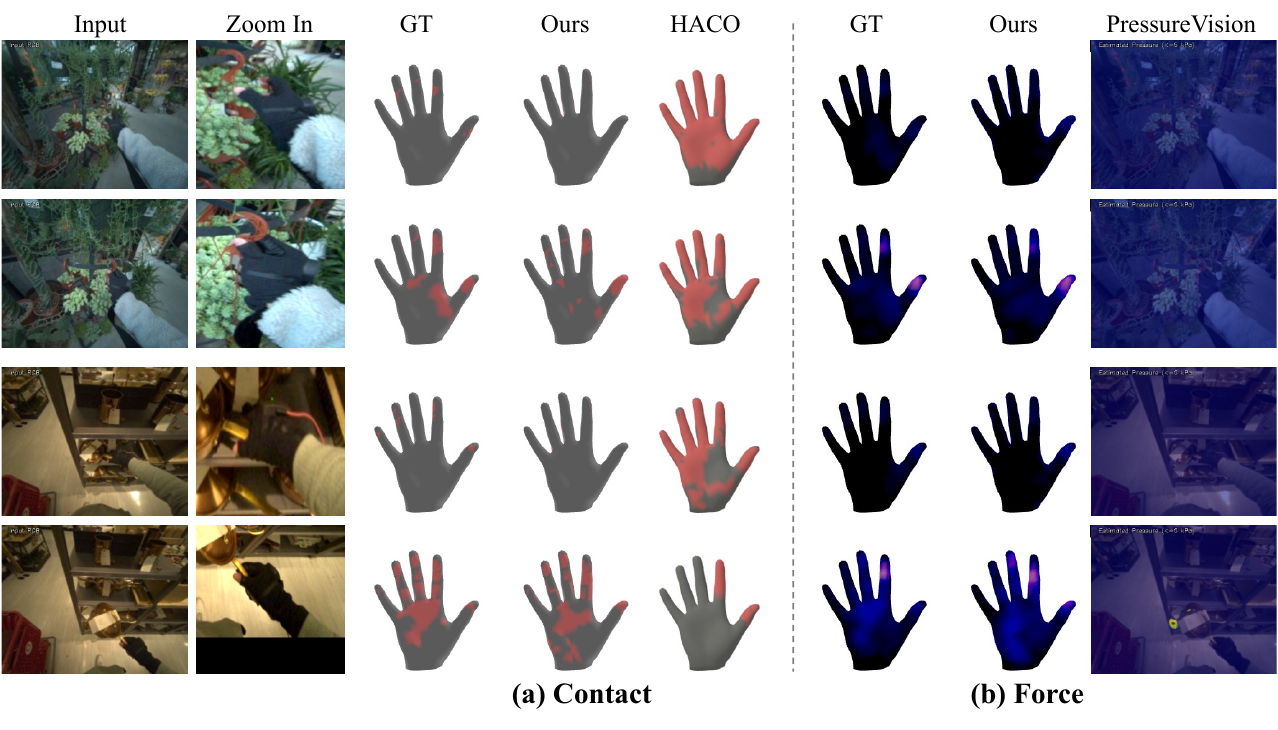}
    \vspace{-15pt}
    \caption{\textbf{Qualitative comparison on OpenTouch.} (a)~Vertex-level contact (b)~Vertex-level pressure}
   \label{fig:comparison_ot}
   \vspace{-12pt}
\end{figure}
\begin{table}[]
\caption{Quantitative comparison of hand-object pressure estimation on OpenTouch dataset.}
\label{table:opentouch}
\centering
\resizebox{1\columnwidth}{!}{
\begin{tabular}{lccc|ccc|cc}
\toprule
 & \multicolumn{3}{c|}{Contact (frame-level)} & \multicolumn{3}{c|}{Contact (vertex-level)} & \multicolumn{2}{c}{Pressure (vertex-level)} \\ \cline{2-9}
 & F1 & Precision & Recall & F1 & Precision & Recall & MAE (kPa) & RMSE (kPa) \\ \midrule
PressureVision \cite{grady2022pressurevision} & 0.164 & 0.767 & 0.092 & 0.013 & 0.239 & 0.007 & 1.55 (5.96 / 0.18) & 4.43 (9.05 / 0.59) \\
PressureVision++ \cite{grady2024pressurevision++} & 0.008 & 0.923 & 0.004 & 0.001 & 0.404 & 0.001 & 1.53 (5.98 / 0.16) & 4.44 (9.07 / 0.59) \\
HACO \cite{jung2025haco} & 0.801 & 0.669 & 1.000 & 0.373 & 0.267 & 0.615 & - & - \\
Ours & 0.888 & 0.824 & 0.964 & 0.680 & 0.691 & 0.669 & 1.23 (3.73 / 0.45) & 3.10 (5.87 / 1.38) \\ \bottomrule
\end{tabular}}
\end{table}

\section{Experiments}
\subsection{Experimental setup}
\textblock{Datasets.}
We train and evaluate our model on a combination of pressure and contact datasets that cover different forms of hand interaction.
For hand-object pressure, we use \textsc{OpenTouch}~\cite{song2025opentouch}, 
which contains egocentric videos of natural hand-object interaction with synchronized tactile measurements from a wearable glove and hand pose annotations. 
We use 1,644 clips / 171,765 frames for training and 190 clips / 17,264 frames for testing.
For hand-surface pressure, we use PressureVisionDB (\textsc{PVDB})~\cite{grady2022pressurevision}, which contains RGB observations synchronized with pressure maps measured on a planar sensor surface; we use 815 sequences / 170{,}720 frames for training and 204 sequences / 33{,}280 frames for testing.
For contact supervision, we additionally use \textsc{DexYCB}~\cite{Chao2021dexycb} (1{,}144 clips / 10{,}786 frames) and \textsc{ARCTIC}~\cite{fan2023arctic} (193 clips / 74{,}705 frames) with distance-based contact labels. 
\textsc{MOW} \cite{cao2021reconstructing} (92 images, evaluation only), which provides single-image hand-object interaction in the wild, is used for image-level contact evaluation following the protocol of HACO~\cite{jung2025haco}.

\textblock{Baselines.}
We compare with prior methods according to their output space and task.
For hand-surface pressure estimation, we compare against PressureVision \cite{grady2022pressurevision} and PressureVision++ \cite{grady2024pressurevision++}, which predict pressure on planar image coordinates.
For dense hand contact estimation, we compare against HACO \cite{jung2025haco}, which predicts MANO vertex-level hand contact from RGB images.
Since HACO does not estimate pressure magnitude, we report only contact metrics for HACO.
For all baselines, we use the official checkpoints released by the 
authors without further fine-tuning.

\textblock{Metrics.}
We report contact and pressure metrics at multiple granularities. \emph{Frame-level contact} F1/Precision/Recall treats each frame as a binary classification (contact present anywhere on the hand vs.\ not). \emph{Vertex-level contact} F1/P/R is computed per-vertex across all frames, measuring spatial localization of contact. For pressure, we report \emph{vertex-level mean absolute error (MAE) and root mean square error (RMSE)} in kPa, computed over vertices with valid pressure annotations ($\Omega_t^{p}$; glove-covered vertices for OpenTouch and all vertices for PVDB). For OpenTouch, we additionally decompose MAE/RMSE into contact\,/\,non-contact regions (numbers in parentheses) to separately assess pressure magnitude estimation and false-positive suppression. For PVDB, we additionally report \emph{pixel-level contact IoU} and \emph{volume IoU} on the pad plane by projecting MANO-vertex predictions back to sensor coordinates, following the PressureVision protocol.

\begin{figure}[t]
\centering

\begin{minipage}[t]{0.57\linewidth}
    \centering
    \includegraphics[trim={441.5pt 0 0 0pt}, clip, width=\linewidth]{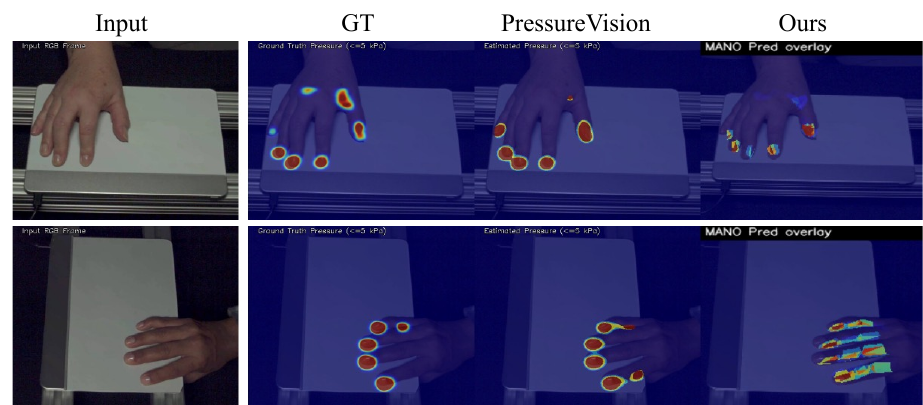}
    \vspace{-14pt}
    \caption{Qualitative comparison on PressureVisionDB. Vertex-level predictions are projected to the image plane.}
    \label{fig:comparison_prv}
\end{minipage}
\hfill
\begin{minipage}[t]{0.40\linewidth}
    \centering
    \includegraphics[width=\linewidth]{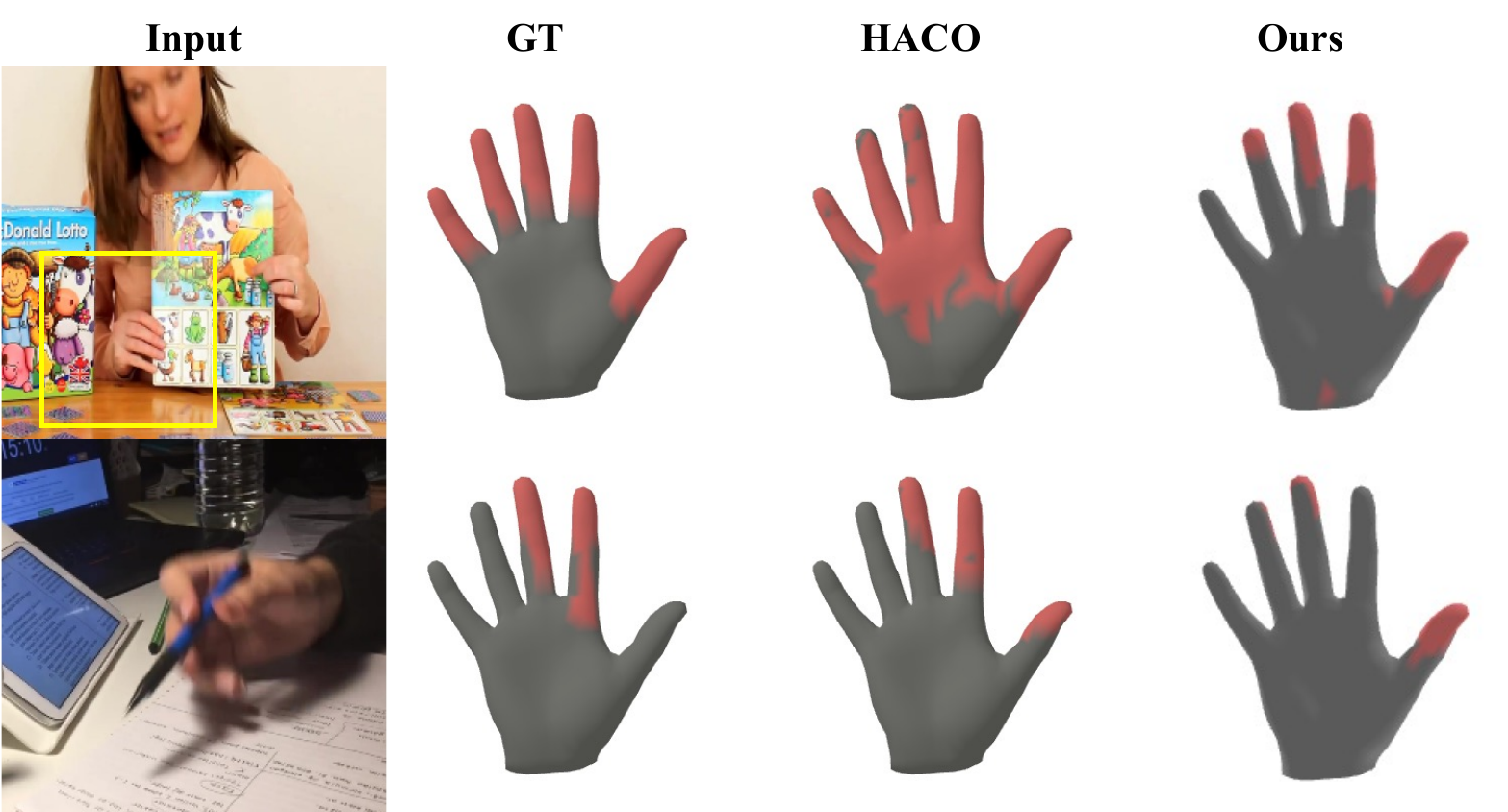}
    \vspace{-14pt}
    \caption{Contact comparison on MOW dataset.}
    \label{fig:mow}
\end{minipage}
\vspace{-10pt}
\end{figure}

\begin{table}[t]
\centering

\begin{minipage}[t]{0.54\linewidth}
    \centering
    \caption{Hand-surface pressure estimation on PressureVisionDB. MAE/RMSE are vertex-level errors in kPa; IoUs are pixel-level.}
    \label{table:pvd}
    \resizebox{\linewidth}{!}{
    \setlength{\tabcolsep}{3pt}
    \begin{tabular}{lcc|ccc}
    \toprule
     & \multicolumn{2}{c|}{Contact} & \multicolumn{3}{c}{Pressure} \\
    \cmidrule(lr){2-3} \cmidrule(lr){4-6}
     & Frame F1 (P / R) & contact IoU & vol IoU & MAE & RMSE \\
    \midrule
    PV \cite{grady2022pressurevision} & 0.939 (0.977 / 0.904) & 0.547 & 0.411 & 0.471 & 3.726 \\
    PV++ \cite{grady2024pressurevision++} & 0.248 (0.963 / 0.143) & 0.022 & 0.021 & 0.248 & 3.025 \\
    Ours & 0.903 (0.901 / 0.906) & 0.326 & 0.210 & 0.303 & 2.073 \\
    \bottomrule
    \end{tabular}}
\end{minipage}
\hfill
\begin{minipage}[t]{0.44\linewidth}
    \centering
    \caption{Dense hand-object contact estimation on MOW dataset.}
    \label{table:mow}
    \resizebox{\linewidth}{!}{
    \setlength{\tabcolsep}{12pt}
    \begin{tabular}{lccc}
    \toprule
     & Precision & Recall & F1 \\
    \midrule
    POSA \cite{hassan2021populating}  & 0.134 & 0.128 & 0.101 \\
    BSTRO \cite{huang2022capturing} & 0.204 & 0.126 & 0.112 \\
    DECO \cite{tripathi2023deco}  & 0.246 & 0.235 & 0.197 \\
    HACO \cite{jung2025haco}  & 0.525 & 0.607 & 0.522 \\
    \midrule
    Ours  & 0.549 & 0.420 & 0.422 \\
    \bottomrule
    \end{tabular}}
\end{minipage}
\vspace{-10pt}
\end{table}

\begin{figure}[t]
    \centering
    \begin{minipage}[c]{0.65\linewidth}
        \centering
        \includegraphics[width=\linewidth]{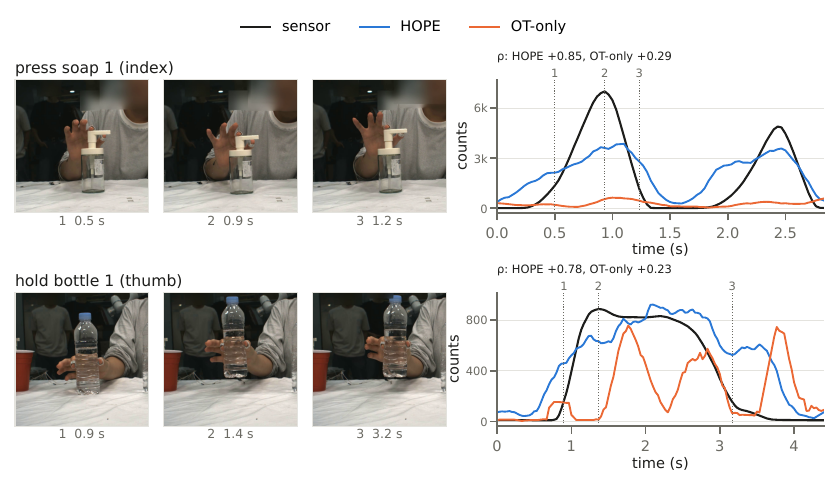}
    \end{minipage}%
    \hfill%
    \begin{minipage}[c]{0.33\linewidth}
        \centering
        \resizebox{\linewidth}{!}{%
        \setlength{\tabcolsep}{5pt}
        \begin{tabular}{@{}lccc@{}}
            \toprule
            & Combined & Thumb & Index \\
            \midrule
            \multicolumn{4}{@{}l}{\emph{Mean Spearman's $\rho$}} \\
            PV & 0.13 & 0.00 & $-0.06$ \\
            PV++ & 0.00 & $-0.06$ & $-0.17$ \\
            OT-only & 0.17 & 0.21 & 0.02 \\
            HOPE & \textbf{0.50} & \textbf{0.42} & \textbf{0.35} \\
            \addlinespace[5pt]
            \multicolumn{4}{@{}l}{\emph{Positive clips ($\rho > 0$)}} \\
            PV & 14/23 & 12/23 & 8/23 \\
            PV++ & 13/23 & 10/23 & 9/23 \\
            OT-only & 15/23 & 17/23 & 12/23 \\
            HOPE & \textbf{23/23} & \textbf{22/23} & \textbf{21/23} \\
            \bottomrule
        \end{tabular}%
        }
    \end{minipage}%

    \par\vspace{-3pt}
    \begin{minipage}{0.65\linewidth}
        \centering
        {(a) Temporal pressure predictions}
    \end{minipage}%
    \hfill%
    \begin{minipage}{0.33\linewidth}
        \centering
        {(b) Quantitative evaluation}
    \end{minipage}

    \caption{
    Near-bare-hand pressure evaluation.
    (a) Selected RGB frames and corresponding fingertip sensor readings (black), compared with HOPE (blue) and OpenTouch-only (orange).
    (b) Mean clip-wise Spearman correlation and number of clips with positive correlation.
    For visualization, predictions are rescaled to sensor counts per clip and model. $\rho$ is computed before rescaling.
    }
    \label{fig:near_barehand}
    \vspace{-6pt}
\end{figure}

\textblock{Implementation details.}
We use DINOv3~\cite{simeoni2025dinov3} ViT-B/16~\cite{dosovitskiy2021vit}  as a visual backbone and HaWoR~\cite{zhang2025hawor} for per-frame MANO parameter estimation; both are kept frozen during training.
Input frames are resized to $224\times224$, yielding $N_p=196$ patch tokens per frame.
Each clip contains $T=10$ frames.
The vertex-anchored transformer uses $L=$ 2 spatio-temporal blocks with 
hidden dimension $d=64$ and 4 attention heads.
We train with AdamW \cite{loshchilov2018adamw} with a cosine learning-rate schedule, learning rate $3{\times}10^{-3}$, batch size 128, for 20 epochs on $1{\times}$RTX~3090.
Loss weights are set to $\lambda_c=1.0$ and $\lambda_p=1.0$; the 
pressure loss is masked out for contact-only HOI samples.
The contact threshold is $\tau_p=1$~kPa and the distance-based 
contact threshold follows HACO at $\delta_c=10$ mm.

\subsection{Comparison}

\textblock{Hand-object pressure on OpenTouch.}
Table~\ref{table:opentouch} reports results on the held-out OpenTouch test set, our main target setting.
Our method achieves the highest vertex-level contact F1 and the lowest pressure errors among methods that predict pressure.
PressureVision predicts pressure in a 2D planar map and cannot transfer meaningfully to MANO vertices, which explains its near-zero vertex-level F1. HACO predicts MANO-vertex contact and achieves high recall, but its precision drops substantially when applied to gloved-hand interaction, indicating a tendency to over-predict contact regions. Qualitative results in \fref{fig:comparison_ot} confirm that our predictions concentrate pressure on physically meaningful contact regions and align spatially with the ground-truth tactile map.

\newcommand{\cmark}{\textcolor{green!70!black}{\checkmark}}
\newcommand{\xmark}{\textcolor{black!40}{$\times$}}

\begin{table}[t]
\centering
\caption{Dataset composition ablation. \textbf{OT}: OpenTouch (pressure), 
\textbf{PV}: PressureVisionDB (pressure), 
\textbf{HOI}: DexYCB + ARCTIC (contact).
We evaluate on OpenTouch (in-domain hand-object pressure), 
PVDB (hand-surface pressure), and ARCTIC (bare-hand contact 
generalization).
}
\label{tab:ablation_dataset}
\resizebox{\columnwidth}{!}{
\small
\setlength{\tabcolsep}{8pt}
\begin{tabular}{ccc|ccc|cc|cc}
\toprule
\multicolumn{3}{c|}{Training data} & \multicolumn{3}{c|}{OpenTouch} & \multicolumn{2}{c|}{PVDB} & \multicolumn{2}{c}{ARCTIC} \\
\cmidrule(lr){1-3} \cmidrule(lr){4-6} \cmidrule(lr){7-8} \cmidrule(lr){9-10}
OT & PV & HOI 
& Frame F1 $\uparrow$ & Vertex F1 $\uparrow$ & MAE $\downarrow$
& Frame F1 $\uparrow$ & MAE $\downarrow$
& Frame F1 $\uparrow$ & Vertex F1 $\uparrow$ \\
\midrule
\checkmark &            &            & 0.886 & 0.669 & 1.239 & 0.729 & 1.094 & 0.825 & 0.157 \\
\checkmark & \checkmark &            & 0.886 & \textbf{0.687} & 1.231 & \textbf{0.910} & \textbf{0.257} & 0.737 & 0.087 \\
\checkmark & \checkmark & \checkmark & \textbf{0.888} & 0.680 & \textbf{1.226} & 0.903 & 0.303 & \textbf{0.954} & \textbf{0.513} \\
\bottomrule
\end{tabular}}
\vspace{-8pt}
\end{table}


\begin{figure}[t]
\centering
\includegraphics[trim={0 0 0 0pt}, width=1\linewidth]{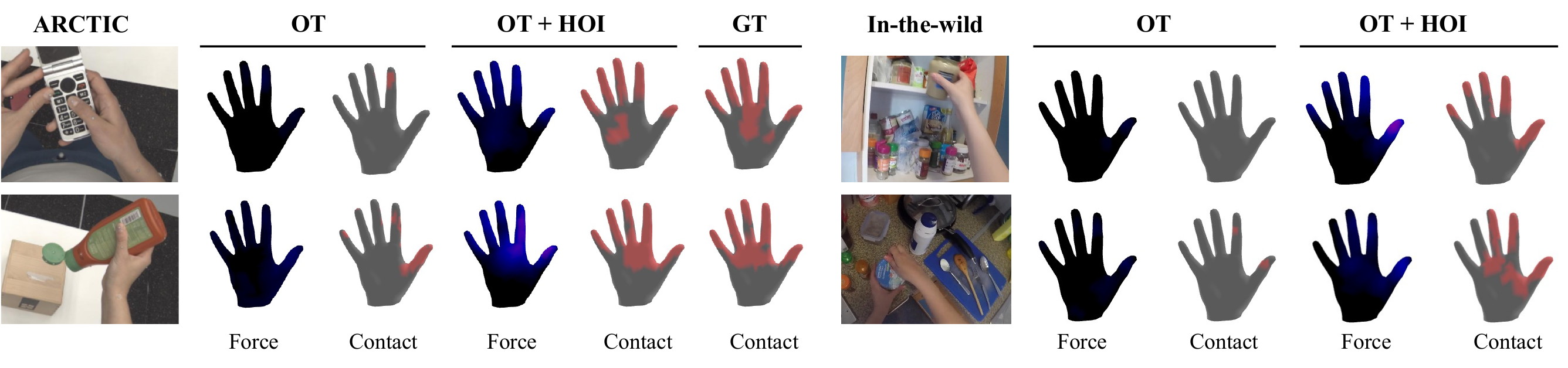}
    \caption{Generalization to bare-hand object interaction. We compare predictions from models trained without (OT) and with (OT + HOI) bare-hand contact supervision. }
   \label{fig:general}
   \vspace{-10pt}
\end{figure}

\textblock{Hand-surface pressure on PVDB.}
We next evaluate on PressureVisionDB to assess whether the same MANO-vertex representation supports planar hand-surface pressure estimation (Table~\ref{table:pvd}).
Our method is competitive with PressureVision on frame-level contact and achieves the lowest pressure RMSE, despite predicting in a hand-centric space rather than the sensor's native 2D image space.
PressureVision retains an advantage on pixel-level contact IoU, as projecting MANO-vertex predictions back to the pad plane introduces hand reconstruction error that planar methods avoid. Overall, these results suggest that the MANO-vertex formulation supports planar-surface pressure estimation while enabling pressure estimation in non-planar settings. Qualitative comparisons are shown in \fref{fig:comparison_prv}.

\textblock{Dense hand-object contact on MOW.}
We also evaluate contact transfer to in-the-wild bare-hand object interaction on MOW (Table~\ref{table:mow}). Our model achieves higher precision than HACO but lower recall, resulting in a lower overall F1.
Qualitative results show that HOPE tends to predict more localized contact regions, consistent with the difference between pressure-derived and proximity-based contact supervision (\fref{fig:mow}). 
Importantly, our model maintains reasonable contact performance while additionally enabling pressure-magnitude prediction, a capability absent in contact-only baselines.

\textblock{Pressure generalization to near-bare-hand interactions.}
To quantitatively assess HOPE's generalization beyond gloved-hand training data, we collect 23 held-out clips (6,310 frames) of everyday object manipulation.
Two pressure sensors are attached to the thumb and index fingertips, leaving most of the hand visually uncovered.
Since the sensors provide uncalibrated readings, we measure temporal agreement using clip-wise Spearman's rank correlation ($\rho$).
We compare HOPE against PressureVision, PressureVision++, and an \emph{OpenTouch-only} variant trained without the additional heterogeneous supervision.
As shown in \fref{fig:near_barehand}, HOPE achieves the highest mean correlation across the overall measure and both fingertips, with positive correlations on nearly all clips.
Compared with OpenTouch-only, HOPE improves the overall correlation from 0.17 to 0.50, outperforming it on 22 of the 23 clips (paired Wilcoxon signed-rank test, $p=2.5\times10^{-5}$).
The temporal examples further illustrate that HOPE more closely follows the measured pressure buildup and release during object manipulation.
These results suggest that heterogeneous supervision improves the transfer of relative pressure dynamics to near-bare-hand interactions.


\begin{table}[]
\centering
\caption{Ablation study on input modalities. We compare variants using different combinations of visual (RGB), hand pose, and temporal information. \textbf{Bold} indicates the best result.}
\label{tab:ablation_input}
\resizebox{\columnwidth}{!}{
\setlength{\tabcolsep}{14pt}
\begin{tabular}{lcccccccc}
\toprule
\multirow{2}{*}{Variant} & \multicolumn{3}{c}{Inputs} & \multirow{2}{*}{Frame F1} & \multicolumn{3}{c}{Vertex-level} & \multirow{2}{*}{MAE (kPa)} \\
\cmidrule(lr){2-4} \cmidrule(lr){6-8}
 & Visual & Pose & Temp. & $\uparrow$ & F1 $\uparrow$ & P $\uparrow$ & R $\uparrow$ & $\downarrow$ \\
\midrule
Pose only           & \xmark & \cmark & \xmark & 0.8337 & 0.6161 & 0.6341 & 0.5991 & 1.4472 \\
RGB only            & \cmark & \xmark & \xmark & 0.8567 & 0.6363 & 0.6480 & 0.6250 & 1.3601 \\
RGB + Pose          & \cmark & \cmark & \xmark & 0.8656 & 0.6566 & 0.6839 & 0.6314 & 1.2947 \\ \hline
Pose sequence       & \xmark & \cmark & \cmark & 0.8501 & 0.6132 & 0.6756 & 0.5614 & 1.3969 \\
Video               & \cmark & \xmark & \cmark & 0.8822 & 0.6658 & 0.6758 & 0.6561 & 1.2873 \\
\midrule
\textbf{Full (Ours)} & \cmark & \cmark & \cmark & \textbf{0.8884} & \textbf{0.6795} & \textbf{0.6905} & \textbf{0.6689} & \textbf{1.2261} \\
\bottomrule
\end{tabular}}
\vspace{-10pt}
\end{table}


\begin{table}[t]
\centering
\caption{Architecture ablations. (a) Comparison of vertex-token and parameter-token representations. (b) Retraining with different hand reconstructors and visual backbones.}
\label{tab:architecture_ablation}
\begin{minipage}[t]{0.5\linewidth}
\centering
\resizebox{\linewidth}{!}{
\setlength{\tabcolsep}{10pt}
\begin{tabular}{lcc}
\toprule
Representation & Vertex F1 $\uparrow$ & MAE $\downarrow$ \\
\midrule
Parameter tokens + LBS & 0.634 & 1.237 \\
Vertex tokens (Ours) & \textbf{0.680} & \textbf{1.226} \\
\bottomrule
\end{tabular}}
\par\smallskip
{\small (a) Token representation}
\end{minipage}
\hfill
\begin{minipage}[t]{0.48\linewidth}
\centering
\resizebox{\linewidth}{!}{
\setlength{\tabcolsep}{10pt}
\begin{tabular}{lcc}
\toprule
Reconstructor + backbone & Vertex F1 $\uparrow$ & MAE $\downarrow$ \\
\midrule
HaWoR + DINOv3 (default) & 0.680 & 1.226 \\
WiLoR + DINOv3 & 0.661 & 1.280 \\
HaWoR + HaWoR ViT-H & 0.679 & 1.178 \\
\bottomrule
\end{tabular}}
\par\smallskip
{\small (b) Component replacement}
\end{minipage}
\end{table}

\begin{table*}[]
\centering
\caption{Robustness to test-time MANO perturbations on OpenTouch. Each perturbation is drawn once per clip and applied to every frame. The default model reaches 0.680 vertex F1 and 1.226 kPa MAE. We report the change ($\Delta$) from the default.}
\label{tab:robustness}
\resizebox{\textwidth}{!}{
\setlength{\tabcolsep}{4pt}
\begin{tabular}{l|cccc|cccc|cccc|ccc}
\toprule
 & \multicolumn{4}{c|}{Global trans.\ (mm)} & \multicolumn{4}{c|}{Global orient.\ ($^\circ$)} & \multicolumn{4}{c|}{Joint pose ($^\circ$)} & \multicolumn{3}{c}{Shape (std)} \\
 & 10 & 20 & 50 & 100 & 5 & 10 & 20 & 40 & 2 & 5 & 10 & 20 & 0.5 & 1 & 2 \\
\midrule
$\Delta$ Vertex F1 $\uparrow$ & $+$0.0001 & $+$0.0002 & $+$0.0003 & $-$0.0004 & 0.0000 & $-$0.0001 & $-$0.0006 & $-$0.0040 & $-$0.0007 & $-$0.0024 & $-$0.0065 & $-$0.0172 & $-$0.0010 & $-$0.0019 & $-$0.0046 \\
$\Delta$ MAE (kPa) $\downarrow$ & $-$0.0001 & $-$0.0002 & $-$0.0005 & $-$0.0009 & $-$0.0001 & $+$0.0006 & $+$0.0043 & $+$0.0195 & $+$0.0013 & $+$0.0047 & $+$0.0131 & $+$0.0336 & $+$0.0004 & $+$0.0014 & $+$0.0049 \\
\bottomrule
\end{tabular}}
\vspace{-8pt}
\end{table*}

\subsection{Analysis}
\textblock{Effect of training data composition.}
We study how each training source contributes to in-domain performance and cross-domain generalization (Table~\ref{tab:ablation_dataset}). Training on OpenTouch alone (OT) already provides strong hand-object pressure performance on the OpenTouch test set, but transfers poorly to the planar PVDB setting and gives only modest contact accuracy on bare-hand ARCTIC interactions. Adding PVDB pressure supervision (OT + PV) substantially improves planar hand-surface pressure estimation while leaving in-domain OpenTouch performance largely unchanged, indicating that the unified MANO-vertex representation absorbs heterogeneous pressure sources without interference. 
Since both OT and PV expose the model only to gloved or planar-surface interactions, contact transfer to ARCTIC remains limited.
Incorporating distance-based contact from HOI datasets (OT + PV + HOI) substantially improves bare-hand contact localization on ARCTIC, while slightly increasing pressure error on PVDB.
These results highlight the complementary role of contact-only supervision in extending pressure estimation to diverse bare-hand interactions. Qualitative comparisons in \fref{fig:general} show the corresponding effect on bare-hand transfer.

\textblock{Input modality ablation.}
We ablate the three input modalities of the vertex-anchored transformer---visual features, hand pose, and temporal context---in Table~\ref{tab:ablation_input}: RGB is removed by skipping the vertex-to-patch cross-attention, pose by replacing the posed MANO vertices with the canonical-pose template (leaving only the per-vertex identity embedding $e_v$), and temporal context by disabling the temporal attention layer. Pose-only and RGB-only variants achieve comparable performance, and combining the two consistently improves over either alone, indicating that hand articulation and visual evidence carry complementary, non-redundant signals.
Temporal context further improves the RGB-based variants, supporting the importance of modeling pressure evolution over time.
The full model achieves the best performance across the reported contact and pressure metrics.

\textblock{Vertex token ablation.}
To assess the effectiveness of vertex-level reasoning, we compare our vertex-token representation with a compact alternative using 18 MANO parameter tokens instead of 778 vertex tokens.
In this variant, spatial and temporal attention operate on parameter tokens, whose features are subsequently expanded to 778 vertex features using the fixed linear blend skinning (LBS) weights of MANO.
As shown in Table~\ref{tab:architecture_ablation}(a), vertex tokens improve contact F1 from 0.634 to 0.680, with slightly lower pressure MAE.
These results support our vertex-anchored design, where vertex-specific representations are maintained throughout the network for fine-grained reasoning over the hand surface.

\textblock{Robustness to hand reconstruction.}
Since HOPE relies on MANO estimates from an off-the-shelf reconstructor, we assess its sensitivity to reconstruction errors (Table~\ref{tab:robustness}). 
Injecting random noise into the estimated MANO parameters at test time causes limited degradation under moderate perturbations, with the largest F1 drop of 0.017 under 20$^\circ$ per-joint pose noise. The model is trained on estimated MANO parameters, so it already sees reconstruction error during training and learns to tolerate it. 
Replacing HaWoR with the single-image reconstructor WiLoR~\cite{potamias2025wilor} slightly reduces contact F1 (0.680 to 0.661) and increases pressure MAE (1.226 to 1.280 kPa), suggesting that reliable hand reconstruction can benefit downstream pressure estimation.
Separately, replacing DINOv3 with HaWoR's ViT-H features achieves comparable contact F1 (0.679) and lower pressure MAE (1.178 kPa), indicating that VertexFormer is not tied to a particular visual backbone (Table~\ref{tab:architecture_ablation}(b)).




\section{Discussion and conclusion}

We presented HOPE, a framework for predicting per-vertex normal pressure and contact from monocular videos.
By lifting tactile-glove pressure, planar-sensor pressure, and distance-based contact annotations into a shared MANO vertex space, HOPE enables joint learning from heterogeneous supervision.
Our experiments show that contact-only HOI supervision improves contact localization in bare-hand interactions, while the near-bare-hand evaluation further demonstrates improved agreement with measured temporal pressure variations.
These results support the effectiveness of combining complementary pressure and contact supervision for hand-object pressure estimation beyond instrumented settings.

\textbf{Limitations.}
HOPE relies on an off-the-shelf hand reconstructor and may inherit its errors under severe occlusion or challenging viewpoints, although our experiments show robustness to moderate reconstruction noise.
Our training supervision also has limitations in sensing coverage and label consistency: tactile measurements cover only instrumented regions, while distance-based contact annotations do not necessarily correspond to physically loaded contact.
Furthermore, our near-bare-hand evaluation uses two uncalibrated fingertip sensors, allowing us to assess relative temporal pressure variations but not absolute pressure accuracy in kPa.
Finally, HOPE predicts only normal pressure, leaving tangential forces and full contact wrench estimation for future work.


\begin{ack}
This work was supported by a research grant from the Youlchon Foundation, RLWRLD, and Institute of Engineering Research at Seoul National University, NRF grant funded by the Korean government (MSIT) [No. RS-2022-NR070498 and No. RS-2026-25540546], and IITP grant funded by the Korea government (MSIT) [No. RS-2024-00439854, No. RS-2025-25441838, No. RS-2025-25442338, and No. RS-2021-II211343]. H. Joo is the corresponding author.
\end{ack}

{\small
\bibliographystyle{plain}
\bibliography{egbib}
}
\clearpage

\newpage
\section*{NeurIPS Paper Checklist}

\begin{enumerate}

\item {\bf Claims}
    \item[] Question: Do the main claims made in the abstract and introduction accurately reflect the paper's contributions and scope?
    \item[] Answer:\answerYes{} 
    \item[] Justification: The three contributions stated in the abstract and Sec. 1 — a hand-centric MANO vertex formulation, unified annotation across heterogeneous supervision, and a contact-gated VertexFormer — are supported by Sec. 3.1, Sec. 3.2, and the experimental results in Sec. 4 (Tables 1–5, Figs. 1 and 6).

\item {\bf Limitations}
    \item[] Question: Does the paper discuss the limitations of the work performed by the authors?
    \item[] Answer: \answerYes{} 
    \item[] Justification: Limitations are discussed in Sec. 5.

\item {\bf Theory assumptions and proofs}
    \item[] Question: For each theoretical result, does the paper provide the full set of assumptions and a complete (and correct) proof?
    \item[] Answer: \answerNA{} 
    \item[] Justification: The paper does not include theoretical results.

    \item {\bf Experimental result reproducibility}
    \item[] Question: Does the paper fully disclose all the information needed to reproduce the main experimental results of the paper to the extent that it affects the main claims and/or conclusions of the paper (regardless of whether the code and data are provided or not)?
    \item[] Answer: \answerYes{} 
    \item[] Justification: Sec. 3 fully describes the unified annotation pipeline (tactile-to-vertex mapping, sensel-to-vertex projection, distance-based contact) and the VertexFormer architecture (vertex token encoding, spatial/temporal attention, contact-gated head). Sec. 4.1 reports datasets and splits, backbone choices (DINOv3 ViT-B/16, HaWoR), input resolution, clip length, transformer depth and width, optimizer, learning rate, batch size, training epochs, loss weights, and contact thresholds, which together are sufficient to reproduce the main results.

\item {\bf Open access to data and code}
    \item[] Question: Does the paper provide open access to the data and code, with sufficient instructions to faithfully reproduce the main experimental results, as described in supplemental material?
    \item[] Answer: \answerYes{} 
    \item[] Justification:  We release our implementation and the newly collected near-bare-hand evaluation data (23 clips, 6,310 frames), along with documentation and instructions for reproducing the reported experiments. Existing benchmark datasets are available from their original sources.

\item {\bf Experimental setting/details}
    \item[] Question: Does the paper specify all the training and test details (e.g., data splits, hyperparameters, how they were chosen, type of optimizer) necessary to understand the results?
    \item[] Answer: \answerYes{}
    \item[] Justification:  Sec. 4.1 specifies all training and test details: dataset splits for OpenTouch, PVDB, DexYCB, ARCTIC, and MOW; backbone choices (DINOv3 ViT-B/16 and HaWoR, both frozen); input resolution and patch token count; clip length; transformer depth, hidden dimension, and number of heads; optimizer, learning rate, batch size, and training epochs; loss weights; and contact thresholds.

\item {\bf Experiment statistical significance}
    \item[] Question: Does the paper report error bars suitably and correctly defined or other appropriate information about the statistical significance of the experiments?
    \item[] Answer: \answerNo{} 
    \item[] Justification: Results in Tables 1–5 are reported as single-run numbers (with a fixed seed) without error bars or confidence intervals, due to the computational cost of training the full model across multiple seeds on the combined dataset.

\item {\bf Experiments compute resources}
    \item[] Question: For each experiment, does the paper provide sufficient information on the computer resources (type of compute workers, memory, time of execution) needed to reproduce the experiments?
    \item[] Answer: \answerYes{} 
    \item[] Justification: Sec. 4.1 reports that training is conducted on a single RTX 3090 GPU with the specified optimizer, batch size, and number of epochs. Inference uses frozen DINOv3 ViT-B/16 and HaWoR backbones together with the lightweight VertexFormer head described in Sec. 3.2. It took about ~8 hours for training.
    
\item {\bf Code of ethics}
    \item[] Question: Does the research conducted in the paper conform, in every respect, with the NeurIPS Code of Ethics \url{https://neurips.cc/public/EthicsGuidelines}?
    \item[] Answer: \answerYes{} 
    \item[] Justification: The study uses publicly available research datasets and an additional near-bare-hand evaluation recorded by one of the authors. No external participants were recruited for the additional capture.

\item {\bf Broader impacts}
    \item[] Question: Does the paper discuss both potential positive societal impacts and negative societal impacts of the work performed?
    \item[] Answer: \answerNo{}
    \item[] Justification: The paper does not include a dedicated broader impacts discussion. The proposed method targets hand-object pressure estimation as a perception task, with potential applications in AR/VR haptics, robot learning, and rehabilitation, and inherits the privacy considerations of general egocentric vision.
    
\item {\bf Safeguards}
    \item[] Question: Does the paper describe safeguards that have been put in place for responsible release of data or models that have a high risk for misuse (e.g., pre-trained language models, image generators, or scraped datasets)?
    \item[] Answer: \answerNA{} 
    \item[] Justification: The paper does not release new datasets or pretrained models that pose a high risk of misuse. The method builds on publicly available datasets and existing pretrained backbones.

\item {\bf Licenses for existing assets}
    \item[] Question: Are the creators or original owners of assets (e.g., code, data, models), used in the paper, properly credited and are the license and terms of use explicitly mentioned and properly respected?
    \item[] Answer: \answerYes{} 
    \item[] Justification:  All existing assets used in this work — datasets (OpenTouch, PressureVisionDB, DexYCB, ARCTIC, MOW) and pretrained models (DINOv3, HaWoR) — are properly cited in the references and used in accordance with their original licenses and intended research use.

\item {\bf New assets}
    \item[] Question: Are new assets introduced in the paper well documented and is the documentation provided alongside the assets?
    \item[] Answer:  \answerYes{}
    \item[] Justification: We release the near-bare-hand evaluation data with documentation of the capture setup, data format, evaluation protocol, and usage conditions.

\item {\bf Crowdsourcing and research with human subjects}
    \item[] Question: For crowdsourcing experiments and research with human subjects, does the paper include the full text of instructions given to participants and screenshots, if applicable, as well as details about compensation (if any)? 
    \item[] Answer: \answerYes{} 
    \item[] Justification: The near-bare-hand evaluation was recorded with one of the authors as the sole participant. No external participants were recruited, and no crowdsourcing, separate participant instructions, or compensation were involved.

\item {\bf Institutional review board (IRB) approvals or equivalent for research with human subjects}
    \item[] Question: Does the paper describe potential risks incurred by study participants, whether such risks were disclosed to the subjects, and whether Institutional Review Board (IRB) approvals (or an equivalent approval/review based on the requirements of your country or institution) were obtained?
    \item[] Answer: \answerNA{} 
    \item[] Justification: The near-bare-hand evaluation was self-recorded by one of the authors, without external participants.

\item {\bf Declaration of LLM usage}
    \item[] Question: Does the paper describe the usage of LLMs if it is an important, original, or non-standard component of the core methods in this research? Note that if the LLM is used only for writing, editing, or formatting purposes and does \emph{not} impact the core methodology, scientific rigor, or originality of the research, declaration is not required.
    \item[] Answer: \answerNA{} 
    \item[] Justification: LLMs are not used as an important, original, or non-standard component of the core method.

\end{enumerate}

\end{document}